# Can GPT replace human raters? Validity and reliability of machine-generated norms for metaphors

Veronica Mangiaterra[1], Hamad Al-Azary[2], Chiara Barattieri di San Pietro[1], Paolo Canal[1], Valentina Bambini[1]

1) *Laboratory of Neurolinguistics and Experimental Pragmatics (NEPLab), Department of Humanities and Life Sciences, University School for Advanced Studies IUSS, Pavia, Italy*

2) *Department of Humanities, Social Sciences and Communication, College of Arts and Sciences, Lawrence Technological University, Southfield, MI, USA*

**Abstract**

As Large Language Models (LLMs) are increasingly being used in scientific research, the issue of their trustworthiness becomes crucial. In psycholinguistics, LLMs have been recently employed in automatically augmenting human-rated datasets, with promising results obtained by generating ratings for single words. Yet, performance for ratings of complex items, i.e., metaphors, is still unexplored.

Here, we present the first assessment of the validity and reliability of ratings of metaphors on familiarity, comprehensibility, and imageability, generated by three GPT models for a total of 687 items gathered from the Italian *Figurative Archive* and three English studies. We performed a thorough validation in terms of both alignment with human data and ability to predict behavioral and electrophysiological responses.

We found that machine-generated ratings positively correlated with human-generated ones. Familiarity ratings reached moderate-to-strong correlations for both English and Italian metaphors, although correlations weakened for metaphors with high sensorimotor load. Imageability showed moderate correlations in English and moderate-to-strong in Italian. Comprehensibility for English metaphors exhibited the strongest correlations. Overall, larger models outperformed smaller ones and greater human-model misalignment emerged with familiarity and imageability. Machine-generated ratings significantly predicted response times and the EEG amplitude, with a strength comparable to human ratings. Moreover, GPT ratings obtained across independent sessions were highly stable.

We conclude that GPT, especially larger models, can validly and reliably replace – or augment - human subjects in rating metaphor properties. Yet, LLMs align worse with humans when dealing with conventionality and multimodal aspects of metaphorical meaning, calling for careful consideration of the nature of stimuli.

1. **Introduction**

Large Language Models (LLMs) have gained popularity in the last few years, especially after the release of ChatGPT. Their outstanding abilities in producing language in a human-like way led to a number of new research questions within the psycholinguistic community, concerning the impact that LLMs could have on the study of the nature of language. While the more theoretical debate on LLMs' validity as models of human cognition is still ongoing (Bolhuis et al. 2024; Cuskley et al. 2024; Frank and Goodman 2025), a more practical aspect worth exploring is LLMs' role as research tools in the experimental pipeline. A recent survey reported that up to 80% of researchers across diverse fields of study have used LLM-based tools in their research (Liao et al. 2024), calling for a thorough evaluation of the benefits and risks associated with each phase of this integration (Messeri and Crockett 2024; Binz et al. 2025; Charness et al. 2025).

In language sciences and psycholinguistics, integrating LLMs has often resulted in their use to generate ratings for experimental stimuli (Guenther and Cassani 2025; Conde et al. 2025b). Collecting ratings from human participants is time and resource-consuming (with large-scale datasets requiring from 800 to 2000 participants and from 6 to 15 weeks of data collection, see Kuperman et al. 2012; Warriner et al. 2013; Brysbaert et al. 2014), yet nonetheless essential. Indeed, as demonstrated by various psycholinguistic experiments, the speed and accuracy of word recognition is affected by several linguistic properties. A frequent word like *cat* is processed faster than a less frequent word like *opal* (Brysbaert et al. 2018), and a concrete word like *bookcase* is processed faster than an abstract word like *knowledge* (Barber et al. 2013). Consequently, many normed datasets of words along a variety of dimensions were created (Stadthagen-Gonzalez and Davis 2006; Brysbaert et al. 2014; Scott et al. 2019; Lynott et al. 2020), allowing the scientific community to rely on pre-collected norms, with benefits from the perspective of reproducibility and reuse of existing resources. With the advent of LLMs, researchers explored the possibility of automatically generating ratings by directly prompting the models using natural language (Trott 2024a; Martínez et al. 2024b, a; Brysbaert et al. 2024; Xu et al. 2025; Kewenig et al. 2025). These studies prompted LLMs to provide ratings for single words or multi-word expressions for dimensions such as concreteness, arousal, and familiarity, and compared them to commonly used human-rated datasets. LLMs, despite exhibiting some biases, overall reported good validity.

The effect of psycholinguistic features on language processing becomes even more complex when moving from the study of single words to multi-word expressions. In this case, we see the effect not only of the characteristics of individual words, but also of the entire expression (Arnon and Snider 2010), with relevant features such as syntactic complexity and cloze probability. For figurative expressions, such as metaphors, other additional features come into play, such as familiarity or imageability. Focusing on metaphors (e.g., expressions such as *Lawyers are sharks*), they are a paradigmatic case of non-literal use of language, where interpretation needs to incorporate contextual aspects to go beyond the literally encoded meaning to derive the intended figurative meaning (Wilson and Carston 2007). The role of psycholinguistic features in metaphor processing has been widely explored across a variety of empirical approaches. One of the most consequential variables is metaphor familiarity, which has been reported to influence comprehension speed, with more familiar metaphors leading to shorter response times (Blasko and Briihl 1997). Also, familiarity was shown to affect brain activity, as shown by studies using both Event-Related Potentials, or ERPs (Coulson 2008; Canal and Bambini 2023) and functional neuroimaging (Schmidt and Seger 2009; Bambini et al. 2011). For instance, familiar metaphors show a reduced negative deflection of ERPs around 400 *ms* after stimulus presentation - the so-called N400 time window, which reflects early semantic processing (Lai et al. 2009). Another relevant dimension is the involvement of sensorimotor properties, which are known to be activated when processing novel metaphorical meanings (Al-Azary and Katz 2021), with more concrete metaphors eliciting a more negative peak in the N400 window (Canal et al. 2022).

To allow researchers to investigate the patterns of these effects in metaphor processing, several datasets in many languages have been developed (Katz et al. 1988; Bambini et al. 2014; Campbell and Raney 2016; Cardillo et al. 2017; Citron et al. 2020; Huang et al. 2024; Milenković et al. 2024; Bressler et al. 2025), providing up to 1,000 metaphorical expressions, as in the recent Italian *Figurative Archive* (Bressler et al. 2025). However, besides being time consuming (with a number of participants ranging from 60 to 600 subjects, see Katz et al. 1988; Cardillo et al. 2010; Bambini et al. 2014), these datasets remain limited in terms of the number of items compared to the wide range of possible metaphorical sentences, the high variety of syntactic structures in which metaphors can appear (nominal predicative, e.g., *The lawyer is a shark*, predicate, e.g., *He runs away from his problems*, adjectival, e.g., *silky sunsets*), and the many contexts in which they can be embedded. Therefore, it is particularly relevant to explore automatic approaches to generate ratings, which would allow researchers to efficiently and rapidly extract norms for new stimuli suitable for their specific research questions.

Due to their contextualized representations, LLMs have shown good capacity for capturing the context-dependent meaning of figurative expressions, as indicated by their high accuracy (e.g., 78%, reported by Barattieri di San Pietro et al., 2023) when prompted to interpret metaphors (Barattieri di San Pietro et al. 2023; Ichien et al. 2024). This evidence suggests that LLMs could demonstrate acceptable metaphor ratings. Notably, previous studies employed LLMs to generate creativity scores for metaphors and reported strong performance, yet only through fine-tuned models rather than prompting alone (DiStefano et al. 2024). It remains unexplored whether these models can generate ratings via simple prompts in natural language, which is the most accessible means for researchers. Furthermore, a previous study demonstrated that LLMs do not rely on sensorimotor features when producing metaphors (Mangiaterra et al. 2025), highlighting a weakness of models in representing embodied aspects of language, which was also found in metaphor interpretation (Barattieri di San Pietro et al. 2023). Hence, in addition to the question on the accuracy of model performance via simple prompting, it is unclear how LLMs can deal with certain properties of metaphors, particularly with sensorimotor aspects, given that the integration of multimodal sources of information still represents a challenge for AI models (Barattieri di San Pietro et al. 2023; Chakrabarty et al. 2024).

In this work[1], we aimed to extend the study of LLMs as rating generators from the domain of single words to metaphors, testing the accuracy of the models as a tool to complement or substitute human ratings in generating norms for metaphorical stimuli. In particular, we tested validity, intended as the capacity to approximate human performance, and test-retest reliability, defined as the stability of the results over time, of psycholinguistic ratings generated by three recent GPT models (GPT3.5-turbo, GPT4o-mini, and GPT4o - Achiam et al., 2024) via prompting. We also tested the relation of machine-generated ratings with human behavioral and electrophysiological responses, which, to the best of our knowledge, is an unexplored domain within the field of LLMs' ratings, both for single words and more complex expressions. Furthermore, we conducted a separate analysis on the familiarity ratings obtained for subsets of metaphors displaying different embodiment features.

Given LLMs' promising performance for single-word ratings, we expect good validity of machine-generated ratings for metaphors as well. However, we expect shortcomings to emerge for metaphors with high sensorimotor load and for the imageability dimension, as these aspects of language are still challenging for LLMs, both for single words and metaphors (Barattieri di San Pietro et al. 2023; Mangiaterra et al. 2025; Xu et al. 2025). Also, we expected poor test-retest reliability, as GPT models have shown patterns of inconsistency in their responses (Khademi 2023, but see Hackl et al., 2023) for more positive results), with high sensitivity to small changes in prompts (Zhuo et al. 2024).

## 2. Methods

### *2.1. Material (human ratings)*

As a benchmark for machine-generated ratings, we extracted human ratings for three dimensions (familiarity, imageability, and comprehensibility) for 687 metaphors, of which 469 are in Italian from the *Figurative Archive* (Bressler et al. 2025), originally employed in five studies (Bambini et al. 2013, 2014, 2024; Canal et al. 2022; Bressler et al. 2025), and 218 are in English from three original studies (Campbell and Raney 2016; Al-Azary and Buchanan 2017; Cardillo et al. 2017). We also retrieved ratings for 48 anomalous and 168 literal statements in English and for 48 anomalous and 48 literal statements in Italian. All rating studies were acquired from samples of university students, native speakers of the language under consideration. Familiarity ratings were available for seven studies (639 metaphors), imageability ratings for two studies (178 metaphors), and comprehensibility ratings for one study (48 metaphors). Most of the studies collected ratings on a 7-point Likert scale, while Bambini et al. (2013; 2014) employed a 5-point Likert scale and Al-Azary & Buchanan (2017) a 6-point Likert scale. When needed to compute the overall correlations, ratings were standardized to a 7-point Likert scale. A summary of the material is reported in Table 1, with a specification of which datasets (items and ratings) were available online before GPT models' knowledge cutoff (October 2023). More information on the participants of each study can be found in Supplementary Table 1.

*Table 1. Summary of the datasets used in the study.*

| Measure | Definition | Language | Studies | N° item | Form |
|---|---|---|---|---|---|
| Familiarity | Frequency of experience of the expression | English | Campbell & Raney (2016)* Cardillo et al. (2017)* | 170 metaphors 120 literals | (adj) X is (adj) Y *The incriminating files were a poison arrow.* |
| | | Italian | *Figurative Archive*, originally Bambini et al. (2013*, 2014*, 2024); Canal et al. (2022); Bressler et al. (2025) | 469 metaphors 46 anomalous 46 literals | X is Y (Bambini et al. 2013; Canal et al. 2022; Bressler et al. 2025) *Actors are masks* X – Y (Bambini et al. 2024) *Language - bridge* X of Y (Bambini et al. 2014) *Fog of melancholy* |
| Imageability | Ease with which each expression evoked a visual mental image | English | Campbell & Raney (2016)* | 50 metaphors | X is Y *Wound is a fjord* |
| | | Italian | *Figurative Archive*, originally Bambini et al. (2024) | 128 metaphors | X – Y *Language - bridge* |
| Comprehensibility | How suitable or natural the expression is | English | Al-Azary & Buchanan (2017) | 48 metaphors 48 anomalous 48 literals | X is Y *Sarcasm is a knife* |

Note: Items and ratings from studies marked with * were available online before GPT models' knowledge cutoff.

Among metaphors rated for familiarity, 308 could be split into different subsets according to their sensorimotor properties and were used to compare the performance of LLMs for metaphors with varying embodiment features. Specifically, these included 62 metaphors classified as mental, i.e., where the relation between the concepts is based on psychological characteristics, e.g., *I genitori sono scudi* (Eng. Tr.: "Parents are shields") and 62 metaphors classified as physical, i.e., where the relation is based on physical characteristics, e.g., *Certi cantanti sono usignoli* (Eng. Tr.: "Some singers are nightingales"), from Canal et al. 2022); 60 metaphors based on motion words, e.g., *His thoughts were a pendulum*, and 60 metaphors based on auditory words, e.g., *Her chores were a sad tune* from Cardillo et al. (2017); 32 metaphors with topics referring to body parts, e.g., *Quei bicipiti sono sassi* (Eng. Tr.: "Those biceps are stones") and 32 metaphors describing objects, e.g., *Quella casa è un gioiello* (Eng. Tr.: "That house is a jewel") from the IUSS NEPLab MetaBody study (Bressler et al. 2025).

Moreover, we retrieved behavioral responses (response times) for 64 metaphors from the IUSS NEPLab MetaBody study and electrophysiological responses (EEG) for 252 metaphors from Canal et al. (2022) and Bambini et al. (2024), to investigate the validity of machine-generated ratings with respect to human cognitive and neural processing measures. Canal et al. (2022) and Bambini et al. (2024) explored metaphor processing through Event Related Potentials and included mean EEG amplitudes for each metaphor in the N400 time window for frontal and centro-parietal electrodes. The IUSS NEPLab MetaBody study (Bressler et al. 2025) investigated metaphor processing through a sensicality task (participants were asked to say whether the expression made sense in a time-constrained setting) and included response times (RTs) for each metaphor.

## 2.2. Models

We prompted three GPT models (GPT3.5-turbo; GPT4o-mini; GPT4o - OpenAI et al. 2024) through the API and one GPT model (GPT4o-mini, as it was the only one freely accessible at the time of data collection) through the ChatGPT interface. Models were prompted in June 2025.

These models were chosen based on their extensive use in studies concerning both pragmatic abilities in LLMs (Hu et al. 2022; Barattieri di San Pietro et al. 2023) and the integration of LLMs in scientific pipelines as raters (Gilardi et al. 2023; Trott 2024a; Martínez et al. 2024b; Brysbaert et al. 2024), as well as for their superior performance, compared to other LLMs, on rating tasks (Xu et al. 2025).

## 2.3. Prompting Procedure

To generate ratings, we prompted the models with the same instructions given to human participants in each original study, with minimal adaptation. Slight modifications with respect to the original instructions were introduced to remove any text referring to practical aspects of the experiments (such as the key to press to continue) and to add the specification to limit the answer to the rating value, given the tendency of the models to provide verbose answers. Building upon previous studies (Gilardi et al. 2023; Trott 2024a), we intentionally avoided using any prompt-engineering techniques tailored specifically to GPT models to ensure both the comparability of responses between GPT models and human participants and the reproducibility of findings for psycholinguistic research. An example of a prompt is provided in Supplementary Table 2. The prompting procedure was performed through both the API and the ChatGPT web interface, in two independent sessions for each model.

### 2.3.1. API parameters

Parameters were set to optimize performance, following previous work. Temperature, the parameter modulating the degree of determinism in models' behavior, was set at 0 to reduce randomness in the output and ensure consistent responses (Binz and Schulz 2023; Kosinski 2024; Xu et al. 2025). In addition to the instructions in the prompt to answer with only the rating value, to further limit verbosity, the maximum token number was set at 1. The number of top-k most likely tokens to return was set at 3. This was done with the aim of computing an overall rating, resulting from the combination of the rating token and the token probability. The overall rating, in addition to providing a more precise estimate (Martínez et al. 2024b), could mirror the continuous nature of human ratings, which were obtained by averaging across participants. So, exploiting the API's possibility to extract the associated log probabilities for the most likely tokens (Hill and Abadkat 2023), we derived the overall rating by weighing each of the three most likely ratings for their log probabilities. For example, for the metaphor "Evolution is a lottery", GPT3.5-turbo provided the top three most likely outputs "2", "3", and "1" with log probabilities of 0.768, 0.195, and 0.037, respectively, resulting in an overall rating of 2.158.

*2.4. Statistical Analysis*

To assess the validity of machine-generated ratings, namely the possibility of approximating the human gold standard, we computed a correlation analysis and a substitution analysis.

First, we computed Spearman correlations between human-generated ratings and machine-generated ratings for all items (687 metaphors, 214 literal statements and 94 anomalous statements) separately for each dimension (familiarity, imageability, and comprehensibility) and for the two languages. Then, we computed separate Spearman correlations for each of the subsets of metaphors characterized by different sensorimotor load: mental and physical in Canal et al. (2022), auditory and motion in Cardillo et al. (2017), and body-related and object-related in the IUSS NEPLab MetaBody study (Bressler et al. 2025).

Second, we tested whether machine-generated metaphor ratings hold the same explanatory power as human-generated ratings in predicting human behavioral and ERP responses as recorded in the three studies with RT and ERP measures. To do so, in line with Trott (2024a), we replicated the statistical analysis of the three original studies (Canal et al. 2022; Bambini et al. 2024; Bressler et al. 2025) and substituted human-generated ratings with machine-generated ratings. Finally, we compared the models' goodness-of-fit in terms of Akaike Information Criterion (Bozdogan 1987) and R squared.

Specifically, to replicate the analysis predicting response times in the IUSS NEPLab MetaBody study (Bressler et al. 2025), we fitted separate Linear Mixed-Effects Models using *lme4* and *lmerTest* packages (Bates et al. 2015) for each of the GPT models and for human familiarity ratings, considering human response times from the original study as the dependent variable and GPT or human-generated ratings of familiarity as continuous predictors. Then, we also considered familiarity ratings for the two subsets of metaphors (body-related and object-related) separately.

To replicate the analysis predicting the N400 amplitude in Bambini et al. (2024), we first fitted separate Linear Mixed-Effects Models using *lme4* and *lmerTest* packages (Bates et al. 2015) for each of the GPT models and human familiarity and imageability ratings, considering the ERP response (both in frontal and centro-parietal electrodes) in the N400 window as the dependent variable and human and GPT-generated metaphor ratings of familiarity and imageability as continuous predictors. Then, we substituted GPT-generated familiarity and imageability separately in a more complex Linear Mixed-Effects Model, used in the original study, which also included a series of other human ratings (i.e., metaphoricity, semantic distance, number and strength of metaphorical interpretation).

To replicate the analysis predicting the N400 amplitude in Canal et al. (2022), we first assessed the effect of metaphor familiarity on the ERP response, by fitting separate Linear Mixed-Effects Models using *lme4* and *lmerTest* packages (Bates et al. 2015) for each of the GPT models and human familiarity ratings, with EEG amplitude as dependent variable and ratings of familiarity as continuous predictor. Then, we substituted GPT-generated familiarity in the Linear Mixed-Effects Models from the original study, which included a series of other ratings and subjects' task scores (i.e., word frequency and subjects' score at two Theory of Mind tasks).

To assess the reliability of machine-generated metaphor ratings, we computed Spearman correlations between machine-generated ratings obtained in two independent sessions (each following the same prompting procedure described in Section 2.3.).

Finally, in line with Trott (2024), we conducted an exploratory analysis to examine where models differ most from human ratings, operationalized as absolute error between human and GPT ratings. We fitted a Linear Mixed-Effects Model using *lme4* and *lmerTest* packages (Bates et al. 2015), with absolute error as the dependent variable and the original human ratings as a predictor in interaction with dimension (familiarity, imageability, and comprehensibility) and GPT model (GPT3.5-turbo, GPT4o-mini, GPT4o), to investigate if the GPT models report larger errors for high (or low) human ratings on a certain dimension, and the potential impact of the model used to elicit the ratings.

All analyses were performed in R (R Core Team 2025).

## 3. Results

### 3.1. Correlation analysis

For metaphor familiarity ratings, we found positive correlations, with coefficients ranging from 0.50 to 0.64 for English and from 0.20 to 0.65 for Italian. The larger model (GPT4o) showed the best performance compared to the other two, exhibiting a strong correlation with humans both for Italian and English metaphors. Results obtained with smaller models (GPT3.5-turbo and GPT4o-mini) were comparable to GPT4o for English, with moderate-to-strong correlations, while falling behind the larger model for Italian, with weak-to-moderate correlations.

In the imageability dimension, moderate correlations were obtained for English metaphors, with coefficients ranging from 0.37 to 0.56, while moderate-to-strong correlations were obtained for Italian metaphors, with coefficients ranging from 0.38 to 0.65.

For comprehensibility ratings, which were available for English metaphors only, we found strong positive correlations between human-generated and machine-generated ones, with coefficients ranging from 0.69 to 0.79.

Table 2 displays all correlations between human and machine-generated ratings for the three dimensions for metaphors (for more details on the results for the single studies, see Supplementary Table 3).

*Table 2. Correlations between human and machine-generated metaphor ratings for the three dimensions and the two languages.*

| Measure | Language | GPT3.5-turbo | GPT4o-mini (API) | GPT4o-mini (ChatGPT) | GPT4o |
|---|---|---|---|---|---|
| Comprehensibility | English | 0.69*** | 0.74*** | 0.78*** | 0.79*** |
| Imageability | English | 0.39** | 0.42** | 0.56*** | 0.37** |
| | Italian | 0.38*** | 0.46*** | 0.45*** | 0.65*** |
| Familiarity | English | 0.64*** | 0.56*** | 0.50* | 0.61*** |
| | Italian | 0.20*** | 0.42*** | 0.20** | 0.65*** |

All models strongly aligned with humans when rating English literal and anomalous statements for comprehensibility (all *r*s > 0.96) and Italian literal and anomalous statements for familiarity (*r*s ranging from 0.82 to 0.92), and they moderately aligned when rating English literal statements for familiarity (*r*s ranging from 0.53 to 0.63). A complete report of correlations for literal and anomalous statements can be found in Supplementary Table 4.

The distribution of machine-generated and human-generated ratings across studies is displayed in Figure 1 (see the Supplementary Figure 1 for density plots of single studies).

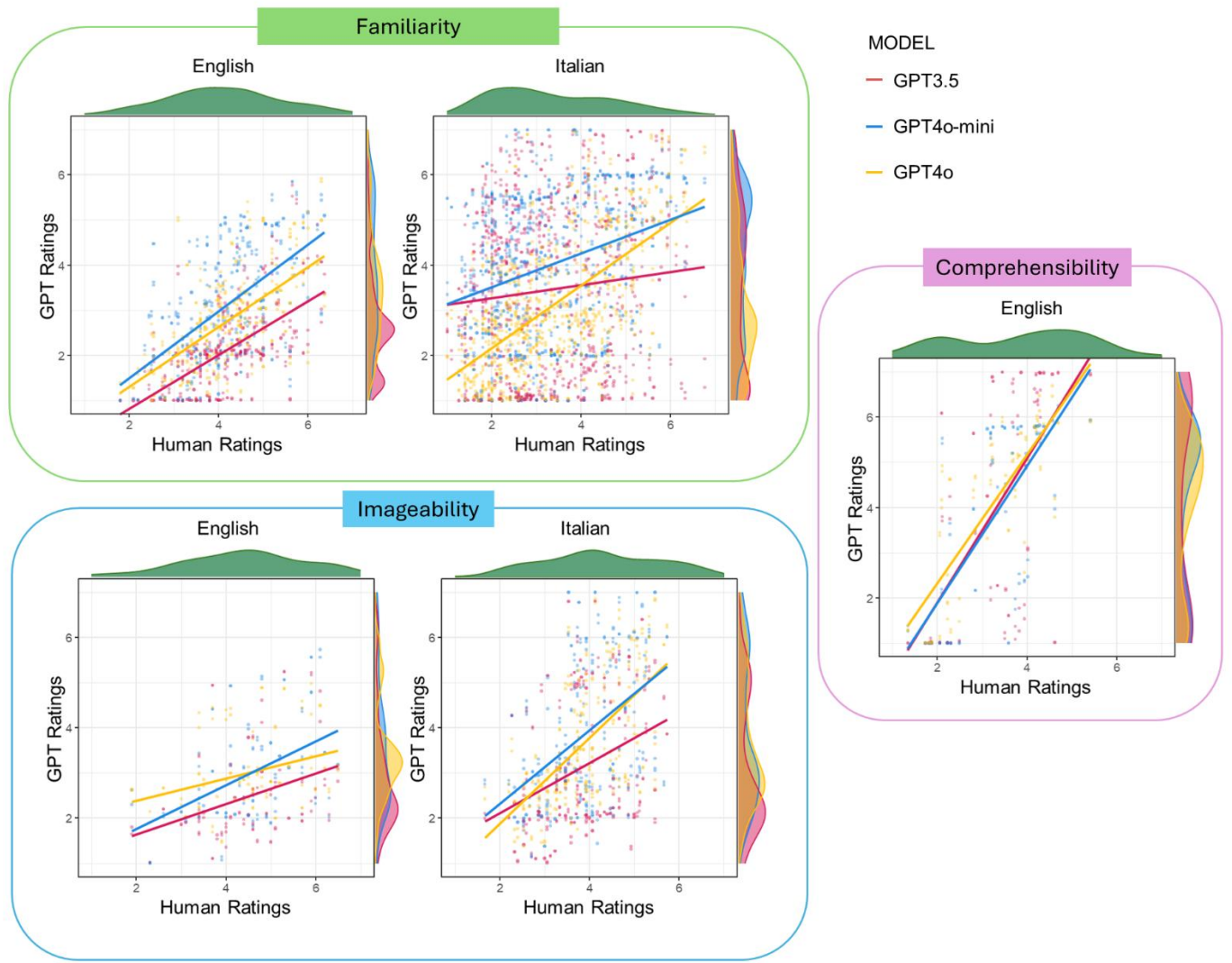

**Figure 1. Distribution of metaphor ratings.** *The figure shows the relationship between GPT ratings (y-axis) and human ratings (x-axis) across three dimensions: Imageability (blue panel), Comprehensibility (purple panel), and Familiarity (green panel), for English and Italian metaphors, with marginal density plots illustrating rating distributions. Regression lines are color-coded by model: GPT3.5-turbo (red), GPT-4o-mini (teal), and GPT-4o (orange).*

### 3.1.1. Correlations between familiarity ratings for metaphors characterized for sensorimotor properties

Correlation analysis (Figure 2) comparing human and machine-generated ratings of familiarity for subsets of metaphors characterized by different sensorimotor properties showed that, for English, ratings for motion metaphors reported strong correlations (GPT3.5: $r = 0.67$; GPT4o-mini: $r = 0.72$; GPT4o: $r = 0.71$), while ratings of auditory metaphors reported moderate to strong correlations (GPT3.5: $r = 0.61$; GPT4o-mini: $r = 0.55$; GPT4o: $r = 0.56$). For Italian, GPT4o-mini and GPT4o showed moderate-to-strong correlations when generating familiarity ratings for object-related metaphors (GPT3.5: *r* = 0.09 (*ns*); GPT4o-mini: *r* = 0.51; GPT4o: *r* = 0.71) and weak-to-strong correlations when rating body-related metaphors (GPT3.5: *r* = -0.16; GPT4o-mini: *r* = 0.37; GPT4o: *r* = 0.65). Familiarity ratings for mental metaphors spanned from weak to strong correlations (GPT3.5: *r* = 0.24; GPT4o-mini: *r* = 0.60; GPT4o: *r* = 0.77), while physical metaphors reported weak-to-moderate performance (GPT3.5: *r* = 0.29; GPT4o-mini: *r* = 0.54; GPT4o: *r* = 0.60). Overall, numerically higher correlations were obtained by metaphors based on motion words, object-related metaphors, and mental metaphors, rather than metaphors based on auditory words, body-related metaphors, and physical metaphors.

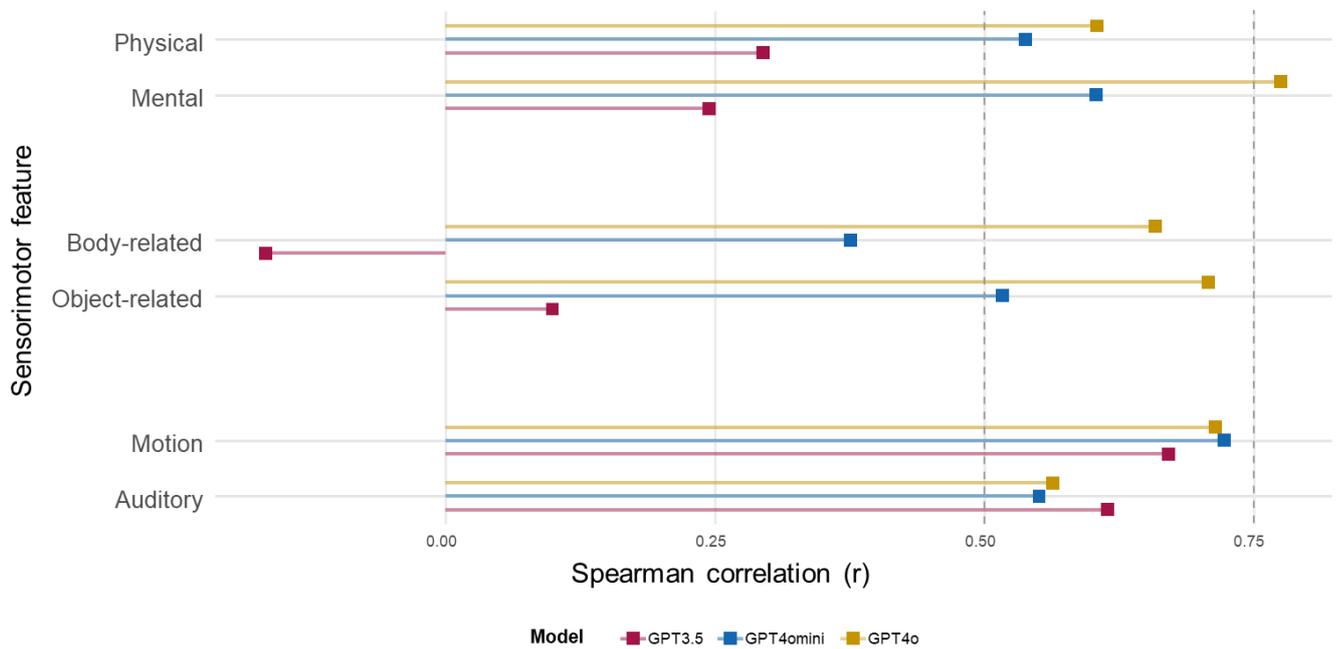

**Figure 2. Correlation between human and GPT ratings of familiarity for metaphors characterized for sensorimotor load.** *Results of the correlation analysis between human-generated and machine-generated familiarity ratings for subsets of metaphors with different types of sensorimotor load (mental and physical from Canal et al. (2022); motion and auditory from Cardillo et al. (2017); object-related and body-related from IUSS NEPLab MetaBody study) for the three GPT models (GPT3.5-turbo, GPT4o-mini, GPT4o).*

### *3.2. Substitution analysis*

#### *3.2.1. Response Times*

Linear Mixed-Effects Models testing the effect of familiarity on response times (data from the IUSS NEPLab MetaBody study) showed a significant effect of machine-generated ratings for both GPT4o-mini prompted through the API ($\beta$ = -0.044, *t* = -5.46, *p* < .001) and GPT4o ($\beta$ = -0.038, *t* = -6.92, *p* < .001), with higher values of familiarity associated with shorter reaction times. This pattern mirrored what we observed in the statistical model with human-generated ratings ($\beta$ = -0.046, *t* = -7.09, *p* <

.001). Explained variance was comparable across models ($R^2 = 0.30$), yet the model with human familiarity had the best goodness of fit (Human: AIC = 1329, GPT4o: AIC = 1331, GPT4o-mini: AIC = 1347). No effect was found for GPT3.5-turbo and GPT4o-mini prompted through the ChatGPT interface (Figure 3a).

Looking at the two subsets of metaphors (body-related and object-related) available in the original study (Bressler et al. 2025), we found that human familiarity and GPT4o familiarity predicted RTs both for body-related (Human: $β = -0.051$, $t = -4.58$, $p < .001$; GPT4o: $β = -0.044$, $t = -4.50$, $p < .001$) and object-related metaphors (Human: $β = -0.041$, $t = -3.34$, $p = .002$; GPT4o: $β = -0.056$, $t = -4.03$, $p < .001$), while GPT4o-mini familiarity only predicted RTs for object-related metaphors ($β = -0.040$, $t = -3.79$, $p < .001$). Again, no effect was found for GPT3.5-turbo and GPT4o-mini prompted through the ChatGPT interface.

*3.2.2. EEG response*

The Linear Mixed-Effects Models examining the effect of familiarity on the N400 amplitude in centro-parietal electrodes (data from Bambini et al., 2024) showed a significant effect of machine-generated ratings (Figure 3b) for GPT3.5-turbo ($β = 0.55$, $t = 2.84$, $p = .005$), GPT4o-mini prompted through the API ($β = 0.40$, $t = 2.06$, $p = .041$) and GPT4o ($β = 0.57$, $t = 3.03$, $p = .003$). No effect was found for GPT4o-mini prompted through the ChatGPT interface ($p = .098$). We found that more familiar metaphors were associated with reduced negativity, as observed for human-generated familiarity ($β = 0.95$, $t = 3.72$, $p < .001$) in the original study (Bambini et al. 2024). All models explained a comparable portion of variance ($R^2 = 0.13$), and AIC comparison indicated that the model with human familiarity provided the best fit (AIC = 9698.41), although differences in AIC across models were relatively modest (GPT4o: AIC = 9702.48).

Human familiarity significantly predicted EEG amplitude in frontal electrodes as well; however, we did not find an effect of machine-generated familiarity for those scalp locations.

The Linear Mixed-Effects Models examining the effect of imageability on EEG amplitude showed no significant effect of machine-generated ratings in the N400 window for either centro-parietal and frontal electrodes, contrasting with a significant effect of human familiarity in both areas (frontal: $β = 0.66$, $t = 2.06$, $p = .04$; centro-parietal: $β = 0.74$, $t = 2.66$, $p < .01$) in the original study (Bambini et al. 2024).

When considering the more complex Linear Mixed-Effects Models used in the original study (Bambini et al. 2024), we found an effect of familiarity for GPT3.5-turbo ($β = 0.42$, $t = 2.53$, $p = .013$) and GPT4o ($β = 0.38$, $t = 1.99$, $p = .049$) in the same direction as the human-generated familiarity ($β$

= 0.83, *t* = 2.28, *p* = .024), again for centro-parietal electrodes only. All models explained a comparable portion of variance ($R^2$ = 0.14), and the model with familiarity by GPT3.5-turbo had the best goodness of fit (AIC = 19714), yet close to the other models (Human: AIC = 19715; GPT4o: AIC = 19716). Familiarity generated with GPT4o-mini did not significantly predict EEG response in this more complex statistical model. Following Trott (2024a), we checked the direction of the effect for this model, and, even if not significant, machine-generated familiarity showed the same direction as human-generated familiarity. Neither human nor machine-generated imageability reached significance in the more complex statistical model.

For the metaphors from Canal et al. (2022), human familiarity did not significantly predict the EEG amplitude in the N400 window in the original study, and, in line with that, we did not find an effect of machine-generated familiarity as well.

To sum up, machine-generated familiarity ratings, especially from GPT4o, can predict N400 amplitude in centro-parietal electrodes, aligning with human ratings, while no machine-generated imageability ratings reported a significant effect, despite human imageability ratings being predictive.

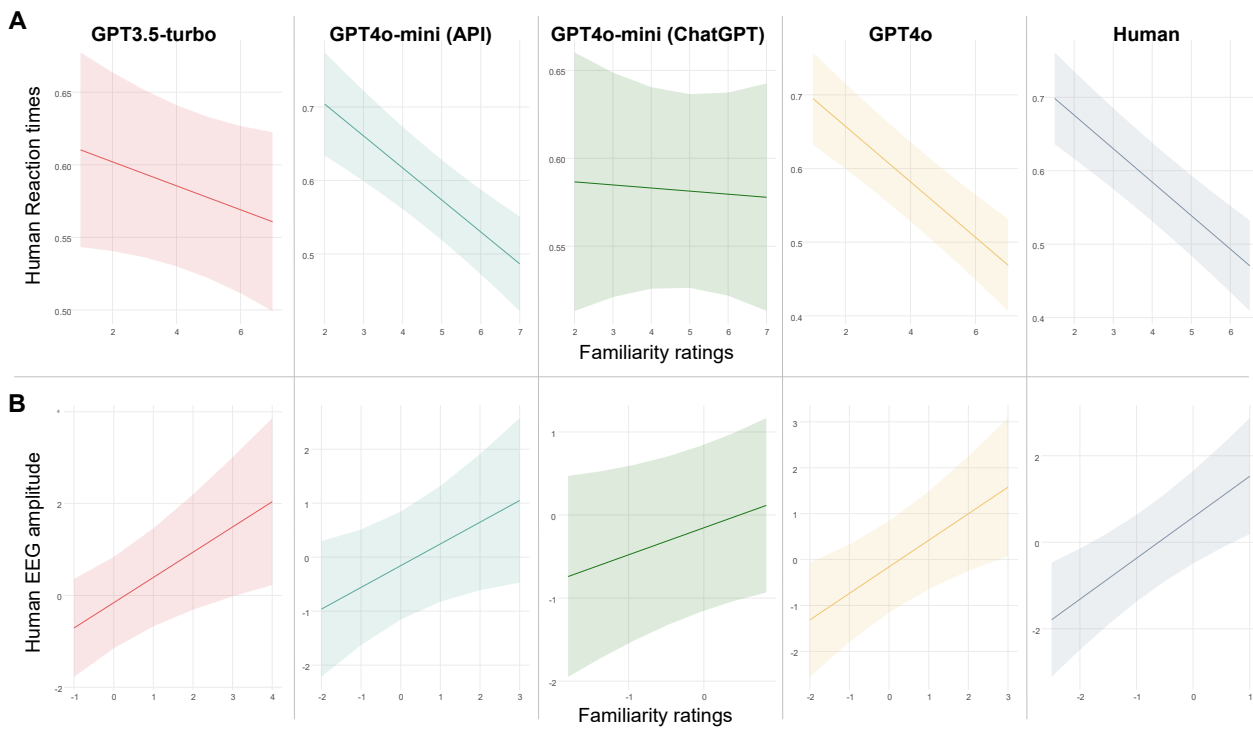

**Figure 3. Results of the substitution analyses**. *Panel A shows the effect of metaphor familiarity ratings generated by GPT3.5-turbo, GPT4o-mini, prompted through the API and ChatGPT, GPT4o, and human participants on Response Times (RTs). Panel B shows the effect of human and machine-generated (GPT3.5-turbo, GPT4o-mini through the API and ChatGPT interface, and GPT4o) metaphor familiarity ratings on the amplitude of the EEG response in the N400 window for centro-parietal electrodes.*

## 3.3. Reliability

The correlations between machine-generated metaphor ratings from two independent sessions showed very high reliability for the three models prompted through the API in all studies, with all correlation coefficients above 0.90 (Table 3). Ratings collected through the ChatGPT interface reported moderate to high reliability, with correlation coefficients ranging from 0.66 (for imageability ratings of English metaphors) to 0.91 (for comprehensibility ratings of English).

Table 3. Correlations between GPT-generated metaphor ratings obtained in two independent sessions for the three dimensions.

| Measure | Language | GPT3.5-turbo | GPT4o-mini (API) | GPT4o-mini (Interface) | GPT4o |
|---|---|---|---|---|---|
| Familiarity | English | 0.99*** | 0.99*** | 0.68*** | 0.98*** |
|  | Italian | 0.98*** | 0.99*** | 0.83*** | 0.98*** |
| Imageability | English | 0.99*** | 0.98*** | 0.66*** | 0.97*** |
|  | Italian | 0.99*** | 0.99*** | 0.67*** | 0.98*** |
| Comprehensibility | English | 0.99*** | 0.99*** | 0.91*** | 0.99*** |

A visual comparison between the reliability of the two prompting methods is shown in Figure 4. For more details on the results for the single studies, see Supplementary Table 5.

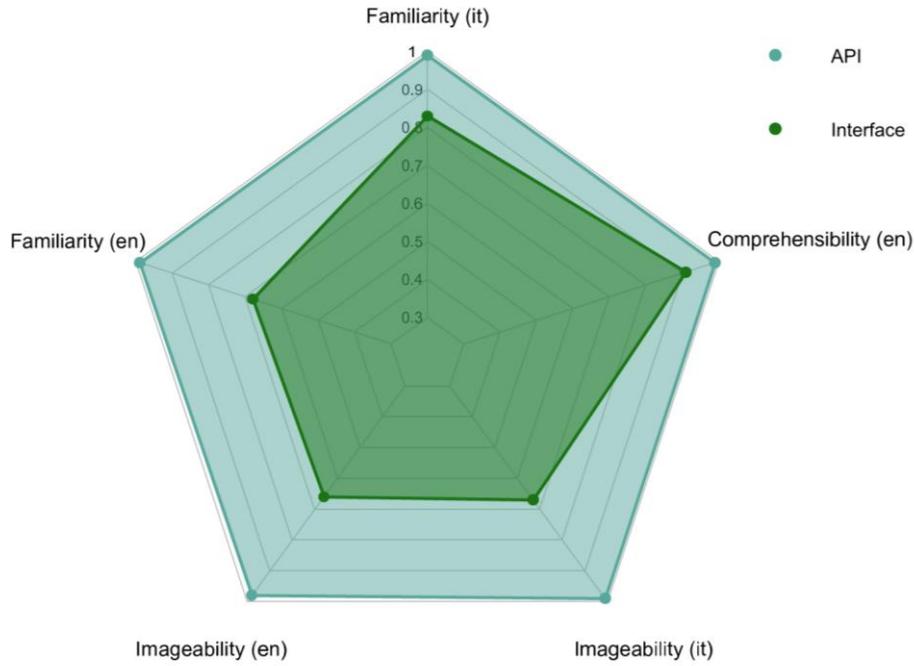

**Figure 4. Reliability comparison between ChatGPT interface and API.** *Visual comparison of the reliability of machine-generated metaphor ratings obtained from GPT4o-mini prompted through the API or the ChatGPT interface.*

## *3.4. Exploratory source of error analysis*

The exploratory analysis of systematic sources of error aimed at testing where human and machine-generated ratings differ. It revealed a main effect of the original human ratings for the metaphors on the absolute error between human and GPT ratings ($β = 0.37$, $t = 13.11$, $p < .001$), with metaphors rated higher by humans associated with higher error. Both the GPT models used to generate ratings and the psycholinguistic dimensions moderated the effect of the original ratings. Specifically, we observed significant interactions between original human ratings and GPT model (GPT4o: $β = –0.14$, $t = -4.28$, $p < .001$; GPT4o-mini: $β = –0.31$, $t = -9.46$, $p < .001$), suggesting that while GPT3.5-turbo aligned less with humans when generating ratings for more familiar, imageable and comprehensible metaphors, more advanced GPT models are less impacted by the original ratings of the metaphor. Trend analyses using the *emmeans* package further supported this finding: the slope of the relationship between human ratings and error was steepest for GPT3.5-turbo ($β = 0.37$) and significantly flatter for GPT4o ($β = 0.23$) and GPT4o-mini ($β = 0.06$).

We also found that the relationship between human ratings and error varied depending on the psycholinguistic dimension, as emerged from the significant interactions between original human ratings and both familiarity ($β = 0.15$, $t = 2.72$, $p < .01$) and imageability ($β = 0.22$, $t = 3.52$, $p < .001$).

Estimated slopes showed that the error increased more steeply with human ratings in the imageability dimension (*β* = 0.33), followed by familiarity (*β* = 0.25), and was the weakest for comprehensibility (*β* = 0.10). Pairwise comparisons confirmed that the slope for comprehensibility was significantly smaller than both familiarity (Δ = –0.15, *p* = .018) and imageability (Δ = –0.22, *p* = .001), whereas familiarity and imageability did not significantly differ (Δ = –0.08, *p* = .24). The GPT models (and especially GPT3.5-turbo) showed lower alignment with humans when generating ratings for high imageable and high familiar metaphors (Figure 5).

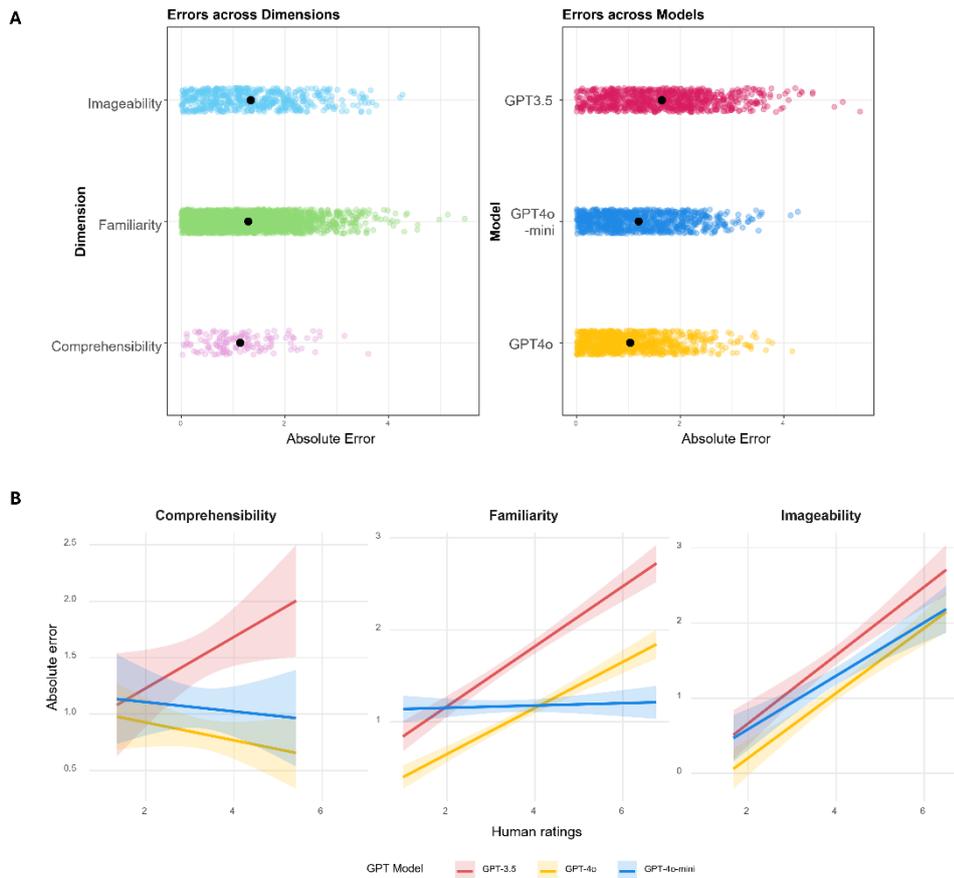

**Figure 5. Absolute error between GPT and human metaphor ratings**. *Panel A shows the distribution of errors for each linguistic feature (Familiarity, Imageability, and Comprehensibility) and each model (GPT-3.5, GPT-4o-mini, GPT-4o). Black points indicate the mean error. Panel B shows absolute error between GPT and human metaphor ratings as a function of original human ratings. Each subpanel shows one dimension (Comprehensibility, Familiarity, Imageability), and each line represents a different GPT model (GPT3.5-turbo, GPT4o, GPT4o-mini).*

## 4. Discussion

The increasing employment of LLMs as annotators or research assistants in the experimental pipeline calls for a systematic assessment of their trustworthiness to build guidelines for researchers that highlight the benefits as well as the potential risks associated with their integration in research. In this study, we extended the line of research investigating the viability of LLMs as raters for norming stimuli in psycholinguistic studies, examining the case of complex figurative expressions such as metaphors. Specifically, we evaluated the validity and reliability of machine-generated ratings across three psycholinguistic dimensions, namely familiarity, imageability, and comprehensibility for metaphorical expressions in English and Italian. To generate ratings, we prompted three GPT models (GPT3.5-turbo, GPT4o-mini, and GPT4o) with the original instructions given to human participants. Models were prompted through both the API and the ChatGPT web interface, allowing for a comparison of prompting settings. To assess performance, we checked whether machine-generated ratings correlate with human ratings and whether they hold the same explanatory power as human ratings in predicting human behavioral and electrophysiological responses. Given the non-converging evidence on the consistency of LLMs' output (Khademi 2023; Hackl et al. 2023), we also examined the stability of ratings across separate sessions.

Our results showed that machine-generated ratings can largely approximate human ratings of metaphors, with high positive correlations emerging for all three dimensions. The larger and most recent model GPT4o reported the highest validity, both for Italian and English, obtaining a strong alignment with human ratings in both languages. Smaller models (GPT3.5-turbo and GPT4o-mini) showed good performance in rating English metaphors, but a sparser pattern of results emerged for Italian metaphors, with two datasets not reporting any association between machine and human ratings. The lower performance of GPT3.5-turbo on Italian is in line with similar findings observed for languages other than English for less recent models, whose training sets were heavily based on English (Rathje et al. 2024).

One issue to be considered when reporting these results is data contamination (Conde et al. 2025b), namely, whether the good performance obtained by GPT models is linked to the presence of the datasets under consideration in their training set. Given that GPT models' training data has not been released, we could not conclude that some of the datasets in our study could be part of it. However, four out of eight datasets (364 metaphors out of 687) were not available online before the models' knowledge cutoff, supporting the soundness of our results despite the possible data contamination issue for a portion of our materials.

The validity of GPT-generated metaphor ratings is further supported by the substitution analyses. Indeed, when put in relation to human behavioral and electrophysiological responses, machine-generated familiarity (except ratings obtained through the ChatGPT web interface) predicted response times and EEG amplitude comparable to human-generated familiarity ratings, demonstrating the possibility of substituting human ratings also in complex statistical analyses to model processing patterns. In the literature, many studies reported significant associations of the measure of lexical probability (or surprisal) from LLMs with neural and behavioral measures of processing costs (de Varda et al. 2023; Michaelov et al. 2024). Even if measures generated via prompting are known to be less reliable than measures derived by accessing internal states of LLMs, such as lexical probability (Hu and Levy 2023), our results support the validity of GPT-generated ratings in terms of ability to capture processing mechanisms, opening the possibility to model behavioral and brain data based on a series of other relevant dimensions – figurative familiarity and comprehensibility - rather than only widely used lexical surprisal.

Contrary to our predictions, machine-generated metaphor ratings reported excellent reliability for models prompted through the API. However, the model prompted through the ChatGPT interface showed a sparser pattern of reliability, highlighting the importance of controlling parameters, such as temperature, to ensure a more stable and deterministic behavior, a step possible only when prompting the models through the API.

While these results confirm the promising use of LLMs as raters reported for single words (Trott 2024a; Brysbaert et al. 2024) and fixed multi-word expressions (Martínez et al. 2024b), a number of weaknesses emerged that deserve special consideration. First, as highlighted by the error analysis, higher misalignment between humans and machines emerged for more familiar metaphors. While humans spanned across all the values in the scale and assigned high ratings to conventional metaphors by treating them as fully acceptable expressions, models assigned lower values to conventional figurative expressions, maintaining a clearer boundary between metaphorical and literal language. This might suggest that models adopt a more rigid distinction between literal and figurative, being less sensitive than humans to the shades of this continuum as a function of conventionality (Bowdle and Gentner 2005; Mashal and Faust 2009; Sperber and Wilson 2012) and more to regularization (Ilievski et al. 2025). The excellent performance obtained by the models for literal and anomalous statements is in harmony with this view, indicating the models' ability in recognizing the two extremes of the scale of sense. Future research could further test this speculation by assessing, for instance, the validity of LLM-generated ratings for highly conventional yet figurative expressions such as idioms (for initial evidence, see O'Reilly et al., 2025).

Perhaps the most relevant limitation of LLMs as metaphor raters regards their ability to capture the embodied aspects of meaning. This emerged both in the analysis on the subset of metaphors with different sensorimotor features and in the error analysis. In the former case, models showed strong performance in rating familiarity for mental metaphors but only moderate performance for physical metaphors, suggesting a lack of perceptual experience hampers a closely human-like representation of meaning. Within concrete metaphors, the models showed strong performance for the motion and the object-related items and moderate-to-strong performance for the auditory and body-related items, showing that when the perceptual features are more represented in the lexicon (see Winter et al., 2018 for the greater prevalence of vision words in the lexicon compared to auditory), models can align better to human representations. In the error analysis, models exhibit low alignment with humans when providing ratings for imageability, where greater error between humans and machines was reported for highly imageable metaphors. Overall, these results are in line with evidence from multiple studies, which found that LLMs have impoverished representations of sensorimotor aspects of language (Conde et al. 2025a; Xu et al. 2025), as they rely on linguistic more than sensorimotor features of words (Mangiaterra et al. 2025; Lee et al. 2025) and provide more accurate interpretations of metaphorical expressions that do not require embodied simulations (Barattieri di San Pietro et al. 2023). This experimental evidence aligns with theoretical claims that identify the lack of grounding as one of the major points of distance between humans and LLMs (Borghi et al. 2023; Chemero 2023, but see Pavlick, 2023 for an alternative perspective).

Beyond these aspects, some other limitations should be considered when using LLMs to generate ratings. First, even if we can simulate the continuous nature of the ratings with an *ad-hoc* setting of hyperparameters, LLMs at this point can only approximate an average human participant, or the *wisdom of the crowd* (Trott 2024b). This does not allow for focus on individual variability, for which the recruitment of human participants is still essential (Qiu et al. 2025). Second, the average human participant that LLMs mimic is representative of the perspective of only a certain demographic. Casola et al. (2024) found that LLMs align with the perspective of young participants, while Martínez et al. (2024a) reported that LLMs align less with children's and extra-European participants' ratings. This is a limitation of existing human-normed datasets, as most participants of rating studies were university students (Bressler et al. 2025). Thus, this evidence indicates that on the one hand, LLMs can approximate existing datasets by aligning with their predominant demographic, but on the other hand, continue to overrepresent certain groups to the disadvantage of less investigated samples of participants (Wang et al. 2025).

Given the limitations above, and capitalizing on the experience gained in this study, in Table 4 we provide a summary of recommendations to guide psycholinguists towards a careful and evidence-based integration of LLMs into their experimental pipelines.

*Table 4. Summary of recommendations for the use of LLMs to generate metaphor ratings.*

| | |
|---|---|
| Models and parameters | - Prefer larger LLMs, such as GPT4o.<br>- Access LLMs through APIs, carefully setting hyperparameters to ensure reliability (e.g., temperature set at 0).<br>- Compute overall scores by incorporating log probabilities to obtain continuous ratings.<br>- Check the performance of LLMs in the language of interest (e.g., check that training data contained text in that language). |
| Type of stimuli | - Prefer non-conventional items.<br>- Prefer items with low sensorimotor load.<br>- If available, generate ratings for anomalous and literal statements as benchmarks. |
| Type of linguistic features | - Prefer dimensions based on occurrence, such as familiarity and comprehensibility, rather than embodiment, such as imageability. |
| Prompt | - Use a prompt as close as possible to human instructions.<br>- Limit models' verbosity by requiring only the rating as output. |

## 5. Conclusions

The integration of large language models (LLMs) into psycholinguistic research, and cognitive science more broadly, has generated considerable debate (Dillion et al. 2023; Abdurahman et al. 2024; Bisbee et al. 2024; Harding et al. 2024) regarding the extent to which human data can be replaced or augmented by artificial intelligence. Following the question posed by Dillion et al. (2023), "Can AI models replace human participants?", we argue that while humans must remain the primary subjects of investigation when studying how metaphorical expression are processed, LLMs can serve as valuable tools to augment human data, particularly in those stages of the experimental pipeline that precede analyses of human processing, such as collecting ratings for stimuli. This could enable the creation of larger and more diverse materials, including in languages beyond English, allowed by larger and multilingual models, and supporting further research on metaphor and human language processing.

**Notes**

1. Preliminary versions of this work were presented at the International Symposium of Psycholinguistics (ISP) 2025, Barcelona, Spain (28th May 2025), the Researching and Applying Conference 2025, Southfield, US (7th August 2025), and AI Rom III, Dresden, Germany (4th September 2025).

**Data availability statement**

All data and materials are available at [10.5281/zenodo.17911909](10.5281/zenodo.17911909)

**Code availability statement**

Code to generate ratings is available at [10.5281/zenodo.17911909](10.5281/zenodo.17911909)

**Competing interests**

The authors declare no competing interests.

**Ethical approval statement**

This article does not report any original studies involving human participants conducted by the authors. The study involves the reuse of human data previously collected in compliance with local designated committees. Data from 7 studies were obtained from publicly available online repositories under a CC-BY license, allowing for data reuse in aggregated form. Data for one study (Al-Azary & Buchanan, 2017) were shared directly by the original author (HA) under a similar license and were used exclusively in aggregated form in compliance with relevant data privacy regulations.

**Informed consent statement**

This article does not report any original studies involving human participants conducted by the authors.

**CRediT Author contributions**

Veronica Mangiaterra: Methodology, Software, Investigation, Data Curation, Formal Analysis, Visualization, Writing - Original Draft

Hamad Al-Azary: Conceptualization, Methodology, Supervision, Writing – Review & Editing

Chiara Barattieri di San Pietro: Methodology, Supervision, Writing – Review & Editing

Paolo Canal: Formal analysis, Writing – Review & Editing

Valentina Bambini: Conceptualization, Methodology, Writing - Original Draft, Supervision, Resources, Funding acquisition

**Acknowledgements**

This work received support from the European Research Council under the EU's Horizon Europe programme, ERC Consolidator Grant "PROcessing MEtaphors: Neurochronometry, Acquisition and Decay, PROMENADE" (GA: 101045733). The content of this article is the sole responsibility of the authors. The European Commission or its services cannot be held responsible for any use that may be made of the information it contains.